\documentclass{article} 
\usepackage{acl}

\usepackage{amsmath,amsfonts,bm}

\def\eqref#1{equation~\ref{#1}}

\def\1{\bm{1}}

\DeclareMathAlphabet{\mathsfit}{\encodingdefault}{\sfdefault}{m}{sl}
\SetMathAlphabet{\mathsfit}{bold}{\encodingdefault}{\sfdefault}{bx}{n}

\usepackage{times}
\usepackage{latexsym}
\usepackage{hyperref}
\usepackage{url}
\usepackage{dsfont}
\usepackage{graphicx}
\usepackage{multirow}
\usepackage{multibib}
\newcites{appendix}{References}

\newtheorem{theorem}{Theorem}[section]

\newtheorem{definition}[theorem]{Definition}

\newcommand{\doctor}{medical professional}
\newcommand{\doctors}{medical professionals}
\newcommand{\approxtext}[1]{\ensuremath{\stackrel{\text{#1}}{\approx}}}

\usepackage[utf8]{inputenc} 
\usepackage[T1]{fontenc}    
\usepackage{hyperref}       
\usepackage{url}            
\usepackage{booktabs}       
\usepackage{amsfonts}       
\usepackage{nicefrac}       
\usepackage{microtype}      
\usepackage{xcolor}         

\title{Injecting knowledge into language generation: a case study in auto-charting after-visit care instructions from medical dialogue}

\author{%
  Maksim Eremeev\thanks{Work done while at Curai} \\
  Elemental Cognition\\
  New York University\\
  \texttt{eremeev@nyu.edu} \\
  \And
  Ilya Valmianski\\
  AuxHealth \\
  \And
  Xavier Amatriain\\
  Curai Health \\
  \And
  Anitha Kannan \\ 
  Curai Health
}

\begin{document}

\maketitle

\begin{abstract} Factual correctness is often the limiting factor in practical applications of natural language generation in high-stakes domains such as healthcare. An essential requirement for maintaining factuality is the ability to deal with rare tokens. This paper focuses on rare tokens that appear in both the source and the reference sequences, and which, when missed during generation, decrease the factual correctness of the output text. For high-stake domains that are also knowledge-rich, we show how to use knowledge to (a) identify which rare tokens that appear in both source and reference are important and (b) uplift their conditional probability. We introduce the ``utilization rate'' that encodes knowledge and serves as a regularizer by maximizing the marginal probability of selected tokens. We present a study in a knowledge-rich domain of healthcare, where we tackle the problem of generating after-visit care instructions based on patient-doctor dialogues. We verify that, in our dataset, specific medical concepts with high utilization rates are underestimated by conventionally trained sequence-to-sequence models. We observe that correcting this with our approach to knowledge injection reduces the uncertainty of the model as well as improves factuality and coherence without negatively impacting fluency. \footnote{Code is available at \url{https://github.com/curai/curai-research/tree/main/careplan-charting}.}

\end{abstract}



\section{Introduction}

Recent advances in language modeling ({\it c.f.}~ \citet{Dong22,10.1613/jair.1.12918} for survey) have enabled applications across multiple domains including education \citep{Shen21}, jurisprudence \citep{Bell2021}, e-commerce \citep{Zhang20, Xiao2021}, and healthcare \citep{pmlr-v158-valmianski21a,pmlr-v158-compton21a, Amanuel22, soap}.

One of the central challenges in deploying these models in-the-wild is that rare words  tend to have underestimated conditional probability during generation \citep{Luong14, chintagunta2021, holtzman2020curious}. However, in high-stakes applications, many of these rare words are semantically important and need to be preserved. For example, some symptoms, diseases, and medications can be both rare and important \citep{mottaghi2020} ({\it e.g.} knowing that the patient is taking warfarin is extremely important, even if the word ``warfarin'' occurs infrequently). 

Prior approaches for handling rare word generation utilize a copy mechanism \citep{see-etal-2017-get, joshi2020dr, xu-etal-2020-self, Choi2021}. This facilitates copying from the source text using a probabilistic switch to decide if the next output token is generated or copied from the input \citep{see-etal-2017-get}. However, it doesn't properly resolve the main challenge: not all rare tokens are important. Only specific rare tokens ({\it e.g.} warfarin) have a high probability of appearing in the reference sequence when found in the source sequence. In cases where the training data does not have enough structure to disambiguate which rare words are essential, the copy mechanism becomes overly extractive \citep{gehrmann-etal-2018-bottom, see-etal-2017-get}.

Also relevant to this paper are previous works that integrate knowledge into language models \citep{duan-etal-2020-pre, liu22psp}. In entity-centric summarization, \citet{keskar19ctrl, liu-chen-2021-controllable} add key phrases to the prompt, which through the self-attention mechanism influence the output distribution. However, for prompts containing rare tokens, self-attention struggles to capture the prompt-reference dependency, and the marginal probability of rare tokens remains underestimated. \citet{joshi2020dr} extends this approach by not only explicitly including the medical concepts in the input sequence, but also adding a related term to the loss function. However, they still find that for rare tokens the model underestimates the conditional probability during generation. 

Finally, dictionary look-up of rare and out-of-vocabulary words has been studied in \citet{yu-etal-2022-dict, ruzzetti-etal-2022-lacking}. However, these papers focus on finding good representations of specific tokens. In this paper, we tackle the problem of uplifting important rare tokens even when a good representation is not available.

We base our work on the premise that {\it specific} rare tokens ({\it e.g.} warfarin) have a high probability of appearing in the reference sequence if they also appear in the source sequence. The main questions we tackle in this paper are the following: {\it How do we know which rare tokens have a propensity to appear in both the source and the reference? How do we encode this information into the model?}



We study our approach in the healthcare setting, for the concrete problem of after-visit care instruction generation from a medical dialog between patient and \doctor. We define the medical concept utilization rate and utilization-rate-aware training objective in \autoref{sec:approach}, discuss the care plan generation problem and data collection in \autoref{sec:careplan}, describe the sequence-to-sequence model setup in \autoref{sec:setup}, and report experimental results in \autoref{sec:experiments}. 

Our contributions are the following:

\begin{enumerate}
    \item We are the first to explicitly focus on identifying and modeling specific rare tokens that appear in both the source and the reference. We call them ``high utilization concepts.''
    \item  We propose a measure of ``utilization rate'' to identify tokens that comprise ``high utilization concepts.''  We use external knowledge to help with this computation as these tokens can be extremely rare.
    \item We introduce a regularization term during training that leverages token utilization rate to uplift the conditional probability of important rare tokens. 
    \item We demonstrate the application of our approach to the concrete task of generating after-visit care instructions from \doctor-patient dialogue.
\end{enumerate}

We observe performance improvement with both automatic metrics and human evaluation with medical experts.


\section{Approach} 
\label{sec:approach}



In many sequence-to-sequence tasks, certain rare concepts have a high probability to appear in the reference sequence ($\textbf{y}$) if they also appear in the source sequence ($\textbf{x}$). 
We call these concepts ``high utilization concepts'' ($c \in C_{\textnormal{HU}}$) and formally define them in \autoref{eq:high_utilization_concepts}. These concepts are comprised of one or more tokens $c=[\nu_0, \nu_1, ...]$. We hypothesize that a source of factuality errors in many sequence-to-sequence tasks is that learned model underestimate the conditional probability of high utilization concepts $\hat{p}(y_i=\nu, |\mathbf{y}_{<i}, \mathbf{x}, \nu \in c, c \in \mathbf{x},  c \in C_{\textnormal{HU}}) < p(...)$, where $\hat{p}$ denotes the model estimated probability and $p$ is the true probability. 

\begin{definition}
[High utilization concepts]
Given a universe of concepts $\mathcal{C}$, the set of high utilization concepts $C_{\textnormal{HU}}$ is defined as
\begin{align}
C_{\textnormal{HU}}=\left\{c \in \mathcal{C}: \frac{p(c\in\mathbf{y}|c\in\mathbf{x})}{p(c\in\mathbf{y})} \gg 1\right\}
\label{eq:high_utilization_concepts}
\end{align} 

\end{definition}

\autoref{eq:high_utilization_concepts} answers the question {\it “How do we know which rare tokens have a propensity to appear in both source and target?”} while at the same time it works for rare tokens. 

This key insight leads us to define two goals for this work: learn to identify high utilization concepts, and build a utilization-rate-aware training objective.

\subsection{Identifying high utilization concepts using externally provided knowledge}
\label{sec:identify-high-util}

\begin{figure*}[h]
    \centering
    \begin{minipage}{0.49\linewidth}
    \centering
    \includegraphics[width=1\textwidth]{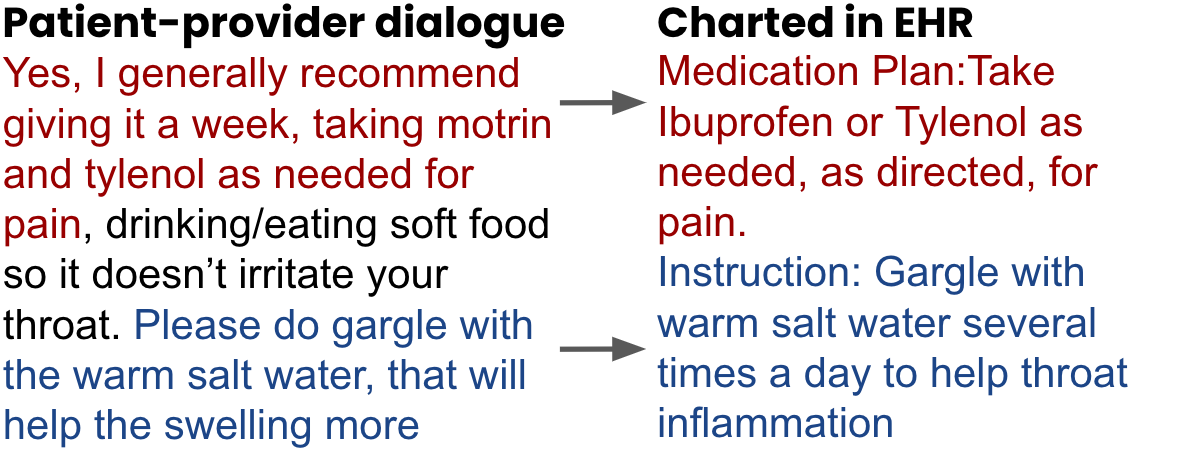}
    (a) A relatively simple-to-chart example with each sentence corresponding to an instruction. Note synonym substitution of ibuprofen for motrin and the addition of timing to the gargling instruction.

    \end{minipage}
    \begin{minipage}{0.49\linewidth}
    \centering
    \includegraphics[width=1.\linewidth]{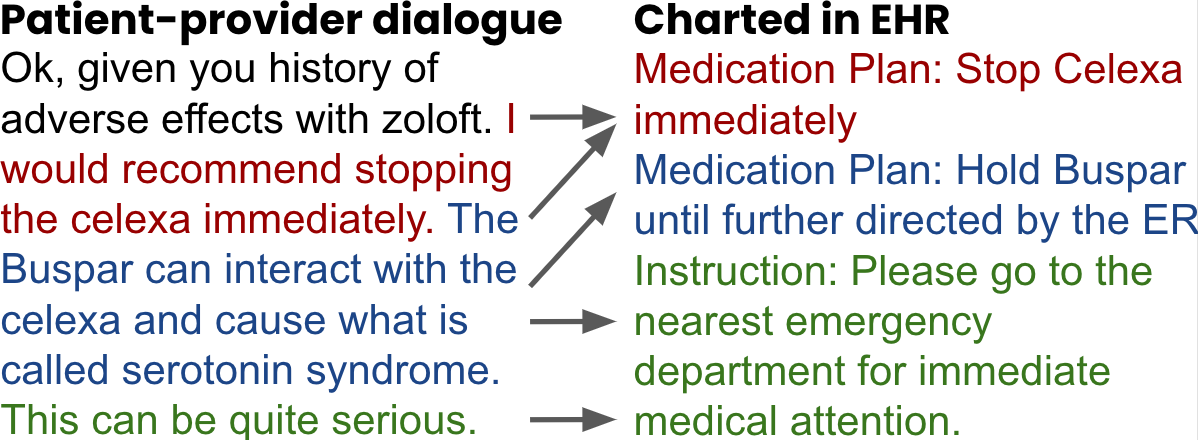}
    (b) A difficult-to-chart example with incomplete information and multiple dialogue sentences contributing to a single instruction.
    \vspace{2em}
    \end{minipage}
    \hfill
    
    \caption{Example conversation segments corresponding to care plan and  corresponding instructions. Color represents the highest overlap between the sentence in the dialogue and the instruction. Arrows represent semantic relationship between the dialogue sentence and instruction. Note that these relationships between the dialog and the instructions are not available in the dataset.}
    \label{fig:careplan_example}
\end{figure*}

The major challenge in identifying high utilization concepts in real datasets is that the concepts we are interested in are present in very few examples. This means that it is hard to directly estimate $p(c\in\mathbf{y}|c\in\mathbf{x})$ and $p(c\in\mathbf{y})$ from \autoref{eq:high_utilization_concepts} due to the high variance. In particular, a frequency-based estimate of probability has an uncertainty proportional to $1/sqrt(N)$ where $N$ is the number of samples for a given concept. However, these rare concepts can still be very impactful to the overall performance of the model. This is because, for a given reference, $\mathbf{y}$, it is unlikely that a \textit{particular} high utilization concept will be present ($\forall c\in C_{\textnormal{HU}}, p(c\in\mathbf{y})\ll1$), but it is also unlikely that \textit{no} high utilization concept will be present ($\prod_{c \in C_{\textnormal{HU}}} p(c\not\in\mathbf{y}) \ll 1$). This is well documented in the medical domain, where medical concepts have a very long-tailed distribution \citep{open-set, mottaghi2020}, yet may appear in almost every relevant sequence. As an illustration, imagine a list of medication instructions. Every instruction may have a different medication so no medication token appears more than once; however, each instruction is rendered useless if it doesn't include the relevant medication ({\it e.g.} see ``Medication Plan'' instructions in \autoref{fig:careplan_example}).

To overcome this challenge, we propose computing what we call ``utilization rate'', $r_{\phi}$, which we define in \autoref{eq:utilization_rate}. This function relies on the concept equivalence class map $\phi: C_\textnormal{sel} \rightarrow \mathcal{E}$ where $C_\textnormal{sel} \subseteq \mathcal{C}$ and $\mathcal{E}$ is a set of equivalence classes. ($\phi$,~$C_{\textnormal{sel}}$,~$\mathcal{E}$) cannot be derived from the data or the model, but instead are provided from an external source of knowledge. If $\phi$ is an identity (id) then $r_{\textnormal{id}}(c_n)= \hat{p}(c_n\in\mathbf{y}|c_n\in\mathbf{x}), (\mathbf{x},\mathbf{y})\in \mathcal{D}$.

\begin{enumerate}
    \item Develop a method for identifying high utilization concepts, $C_{\textnormal{HU}}$ for a dataset  $\mathcal{D}= \{(\textbf{x}^i, \textbf{y}^i)\}_{i=1}^N$.
    \item Develop a method for augmenting the training procedure of sequence-to-sequence models to correctly estimate the conditional probability of tokens forming high utilization concepts. 
\end{enumerate}

\begin{definition}[Utilization rate]
The utilization rate of concept $c_n$  is defined as

\begin{align}
\textstyle{
r_{\phi}(c_n) = \frac{\sum_{c\in C_{\textnormal{sel}}} \sum_{j=1}^N\mathbf{1} [c \in \mathbf{x}^j, c \in \mathbf{y}^j, \phi(c)=\phi(c_n)]}{\sum_{c\in C_{\textnormal{sel}}} \sum_{j=1}^N\mathbf{1} [c \in \mathbf{x}^j, \phi(c)=\phi(c_n)]}}
\label{eq:utilization_rate}
\end{align} 
\end{definition}

Here, \autoref{eq:utilization_rate} tries to make the intuition from \autoref{eq:high_utilization_concepts} applicable to a real dataset. We generally cannot compute the lift because for rare words the dataset frequency derived probability estimates are poor.

Note that \autoref{eq:utilization_rate} combines both externally provided knowledge ($\phi$,~$C_{\textnormal{sel}}$,~$\mathcal{E}$) and dataset derived values. This allows us to inject domain-specific information. Because concepts are mapped to equivalence classes, every concept in a particular equivalence class has the same utilization rate. If a concept $c_n \in C_{\textnormal{sel}}$ has marginal probability to appear in the reference sequence that is much lower than $r_{\phi}(c_n)$ then it is a high utilization concept.


\subsection{Utilization-rate-aware seq2seq training}
\label{sec:regularizer}

Our analysis in \autoref{sec:experiments} (see \autoref{fig:rel_util}) shows that conventionally trained seq2seq models underestimate the utilization rate ($r_{\phi}$) for many rare concepts. While we cannot optimize the utilization rate directly, we can optimize the approximate \textbf{marginal probability} $p(\nu|\mathbf{x})$ of a token $\nu$ given a source sequence $\mathbf{x}$, as seen in \autoref{eq:marginal_prob}.

\begin{align}
\begin{split}
    p(\nu|\mathbf{x}) &= \sum_{\mathbf{y}_{<t}} p(\nu|\mathbf{y}_{<t}) p(\mathbf{y}_{<t}) \approx \\ 
    & \approx \sum_{t=1}^{\|\mathbf{y}\|} p(\nu|\mathbf{y}_{<t})p(\mathbf{y}_{<t}) \approxtext{$p(\mathbf{y}_{<t})$ is uniform} \\ 
    &\approx \frac{1}{\|\mathbf{y}\|}\sum_{t=1}^{\|\mathbf{y}\|} p(\nu|\mathbf{y}_{<t})
 \label{eq:marginal_prob}   
\end{split}
\end{align}

Given the source sequence $\mathbf{x}$, the tokens for which we aim to optimize the marginal probability are $\{\nu \in c, c \in \mathbf{x}  \cap C_{\textnormal{HU}}\}$. We define the unweighted utilization loss.
 
\begin{definition}[Unweighted utilization loss]
\begin{align}
    l_u(\mathbf{x}) = &- \frac{1}{\|\{\nu \in c, c \in \mathbf{x}  \cap C_{\textnormal{HU}}\}\|} \times \\ 
    &\times \sum_{\nu \in c, c \in (\mathbf{x}  \cap C_{\textnormal{HU}})}  \log p(\nu|\mathbf{x})
\label{eq:unweighted-utilization}
\end{align}

\end{definition}

However, not all concepts in $C_{\textnormal{HU}}$ are equally likely to appear in the reference given their appearance in the source. To better reflect we also propose a weighted utilization loss where the weight for each token is determined by its utilization rate.

\begin{definition}[Weighted utilization loss]

\begin{align}
     l_{w}(\mathbf{x}) = - \frac{\sum_{\nu \in c, c \in (\mathbf{x}  \cap C_{\textnormal{HU}})}  r_{\phi}(c) \log p(\nu|\mathbf{x})}{\sum_{\nu \in c,c \in (\mathbf{x}  \cap C_{\textnormal{HU}})}  r_{\phi}(c)}
\label{eq:weighted-utilization}
\end{align}

\end{definition}

Note that \autoref{eq:weighted-utilization} directly injects externally provided knowledge through its dependence on $\phi$.

We use utilization loss as a regularization term and augment the objective function. We use $\alpha > 0$ to balance the strength of the regularization: 
\begin{align}
\label{eq:objective}
    l(\mathbf{x}, \mathbf{y}) = l_{\text{nll}}(\mathbf{y}) + \alpha \cdot l_{u\textnormal{ or } w}(\mathbf{x})
\end{align}
where $l_{\text{nll}} = -\sum_{t=1}^{|\mathbf{y}|}\log p(y_t|\mathbf{y}_{<t}, \mathbf{x})$ and $l_{u\textnormal{ or } w}$ is either $l_{u}$ from \autoref{eq:unweighted-utilization} or $l_{w}$ from \autoref{eq:weighted-utilization}.

\section{After-visit care instruction generation: task and data description}
\label{sec:careplan}

\begin{figure*}[ht]
    \centering
    \begin{minipage}{0.49\linewidth}
    \centering
    \includegraphics[width=0.9\linewidth]{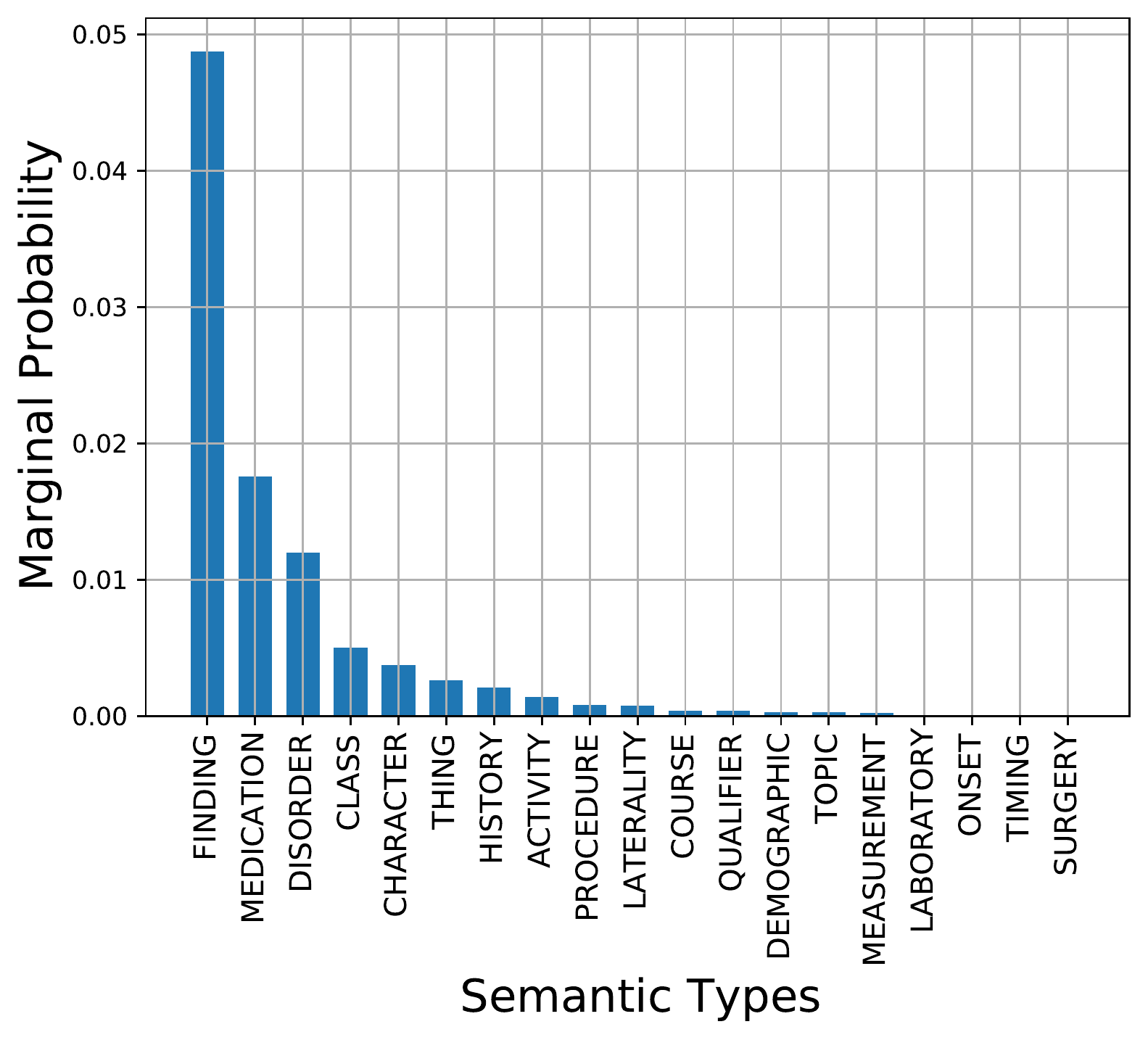}

    (a) Average marginal probability of every semantic type.
    \end{minipage}
    \hfill
    \begin{minipage}{0.49\linewidth}
    \centering
    \includegraphics[width=0.9\linewidth]{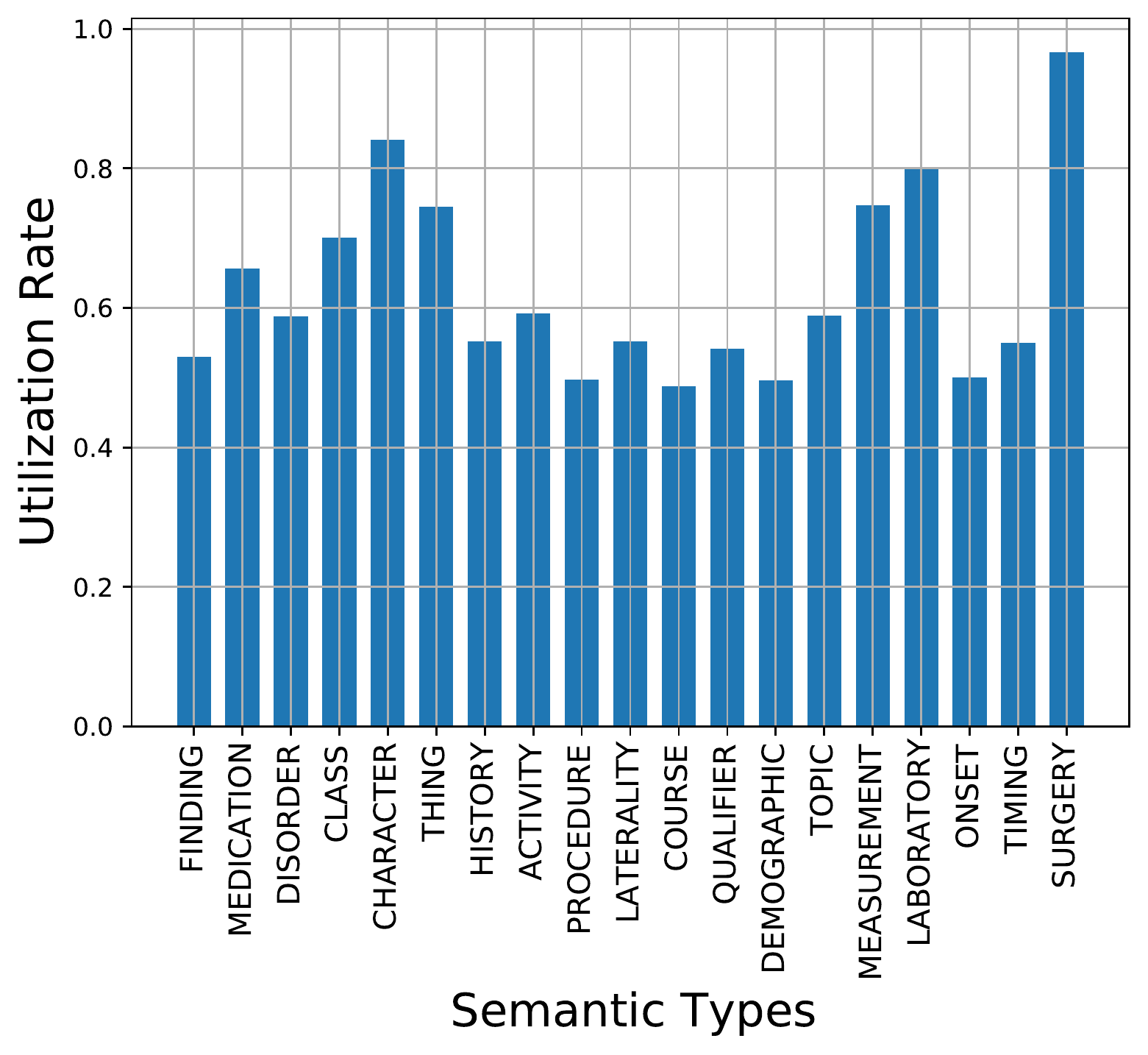}
    
    (b) Semantic type utilization rates.
    \end{minipage}
    \hfill
    
    \caption{Empirical concept marginal probabilities and utilization rates estimated from the dataset.}
    
    \label{fig:concepts_and_utils}
\end{figure*}
   

After-visit care instructions (care plan) are a set of actions (instructions) that a \doctor~writes in the patient's electronic health record (EHR)  as a follow-up to the patient's visit. A care plan often includes a  list of medications with appropriate directions, further medical evaluations, or educational information for preventive care. Before writing the care plan, the \doctor~ discusses it with the patient, and together, they jointly agree on the next course of action.   This joint decision-making implies that most of the necessary information for writing the care plan is already available in the conversation.

In \autoref{fig:careplan_example}, we show two examples. In each example, we present the  (a) segment of the conversational dialog corresponding to provider messages discussing the care plan with the patient and (b) corresponding care plan charted in the EHR. We can see that the  instructions are written in a directive format, using action verbs and often paraphrasings of the corresponding text in the dialogue.  The care plan does not always have all the medical concepts mentioned in the conversation. In the first example, ``serotonin syndrome" and ``Celexa" are rare, but the care plan includes only the latter. We need a  model that is robust to rare medical concepts and can discern which knowledge needs to be carried forward. 

We tackle the problem of taking the relevant section in the conversations corresponding to the care plan as input and automatically derive care plan instructions that the \doctors~can approve. We do not assume access to 1-1 mappings between the sentences in the conversation to the care plan instructions. However, we develop a method to derive a dataset of 1-1 mappings, albeit noisy, which we use for model training.

\paragraph{Dataset construction.} We use a dataset with 14K \doctor-patient encounters collected on a virtual primary care platform. Each encounter has a text-based conversation between the \doctor~ and the patient. We applied an in-house conversation discourse parser to extract only those dialogue turns from the \doctor's corresponding to the care plan discussion. We also have the associated care plans written from the patient's electronic health record for that encounter. On average, each encounter has 9 dialogue turns corresponding to care plans and 4 care plan instructions.

We need a parallel corpus with pairs of dialogue turns and care plan instructions for our model. Getting manual annotations for each encounter would be expensive as it requires expert knowledge. Therefore, we automatically construct a paired dataset, albeit noisily, from the paired encounter level care plan and provider dialog turns. We get sentence-level embeddings for every sentence in each turn and instructions in the care plan and pair those with the highest cosine similarity (We provide additional details in the Supplementary Material). At the end of this, we have 48,000 source-reference pairs, where the source is a sentence in the conversational dialog and reference is the mapped instruction. We randomly sample 3000 pairs for testing, 1000 for validation, and the remaining 44,000 pairs for training.

We use medical concepts from UMLS \cite{UMLS} and in particular SNOMED-CT and RXNorm ontologies. The synonyms are pooled from all ontologies in UMLS that map to the corresponding concept in SNOMED-CT and RXNorm. 

To identify the concepts, we use an in-house lookup-based concept recognizer. It uses a sliding window strategy to find maximal matches of text corresponding to medical concepts and their synonyms. It ignores stop words while doing the match. Finally, it has an agglomeration step that leverages a concept hierarchy. If we have overlapping spans corresponding to two concepts where one is a child of another (eg “lower abdominal pain” and “abdominal pain”) then only the more specific concept is extracted. If two different concepts have a span overlap and are not hierarchically related, then the concept linking is greedily selected with the concept on the left being given priority.

\paragraph{Identifying high utilization concepts.} We limit $C_{\textnormal{sel}}$ to only medical concepts and choose $\phi$ such that it maps them to their SNOMED CT semantic types (which informs our choice of $\mathcal{E}$). In our case study this narrows down 758 unique medical concepts to their 19 semantic types. The marginal probability $p(c \in \mathbf{y})$ for each semantic type $c$ is shown in \autoref{fig:concepts_and_utils}a while the utilization rates are shown in \autoref{fig:concepts_and_utils}b. Comparing them we can see that utilization rates are 10-100x larger than the marginal probabilities. This suggests that all medical concepts are part of high utilization tokens set ($C_{\textnormal{HU}}=C_{\textnormal{sel}}$). It also means that many kinds of medical concepts that are present in the source sequence do not get generated in the output sequence, which drastically hurts medical correctness.
     
\section{Experimental setup}

\begin{figure*}[ht]
    \centering
    \begin{minipage}{0.49\linewidth}
    \centering
    \includegraphics[width=1.\linewidth]{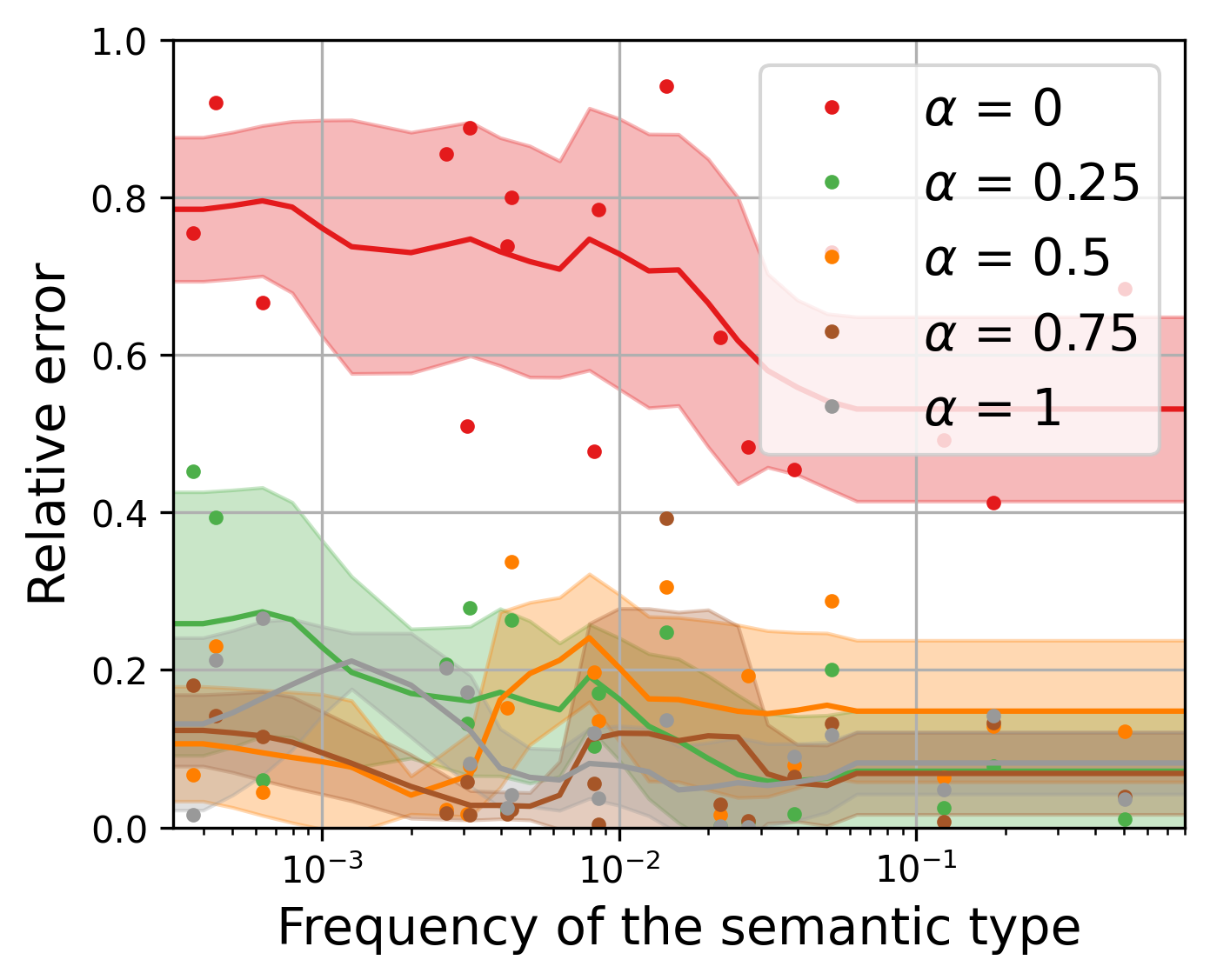}
    
    (a) Relative error in the utilization rate for each regularizer strength $\alpha$. Note that $\alpha=0$ means there is no regularization.
    \end{minipage}
    \hfill
    \begin{minipage}{0.49\linewidth}
    \centering
    \includegraphics[width=\linewidth]{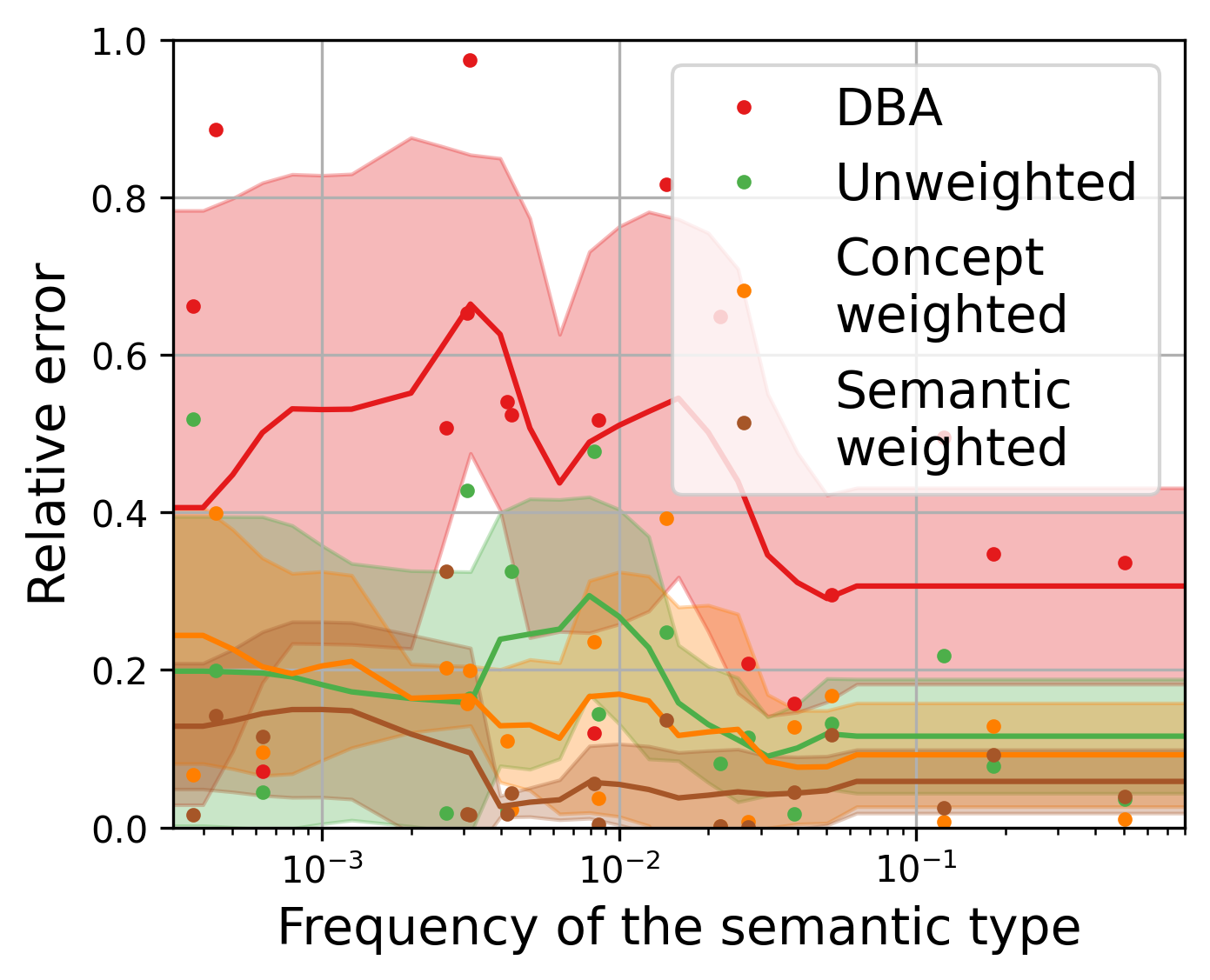}

    (b) Relative error in the utilization rate for each model (and best $\alpha$ for that model).
    \vspace{1em}
    \end{minipage}

    \caption{Relative errors in the utilization rates for different semantic types plotted as a function of the frequency of the semantic type. The trend-line and uncertainty are computed with a linearly interpolated moving average window.}
    \label{fig:rel_util}
\end{figure*}

\begin{figure*}[ht]
    \centering

    \begin{minipage}{0.358\linewidth}
    \centering
    \vspace*{-0.27em}
    \includegraphics[width=1\linewidth]{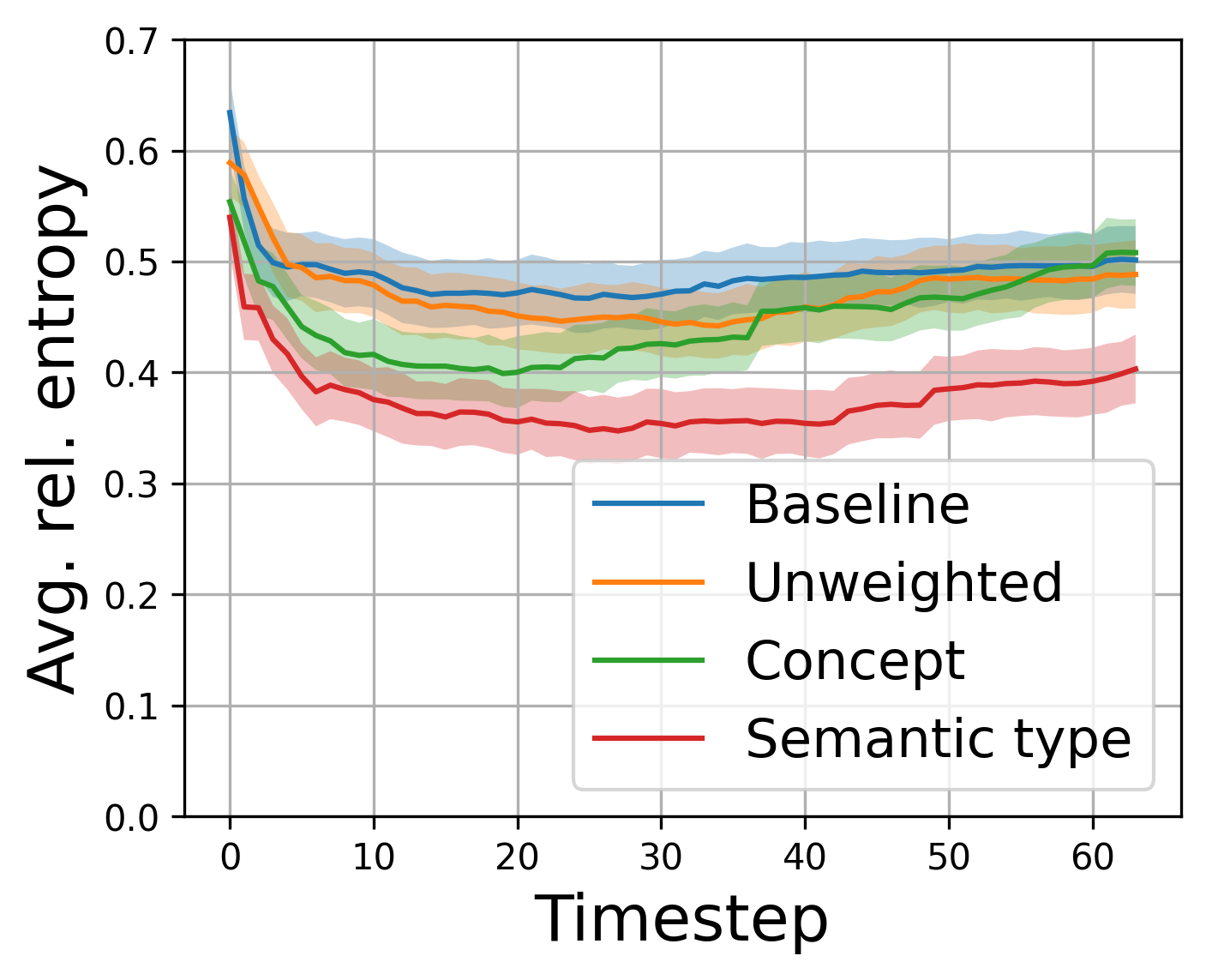}
    
    (a) $\alpha=0.25$
    \end{minipage}
    \hspace{-30pt}
    \hfill
    \begin{minipage}{0.32\linewidth}
    \centering
    \includegraphics[width=\linewidth]{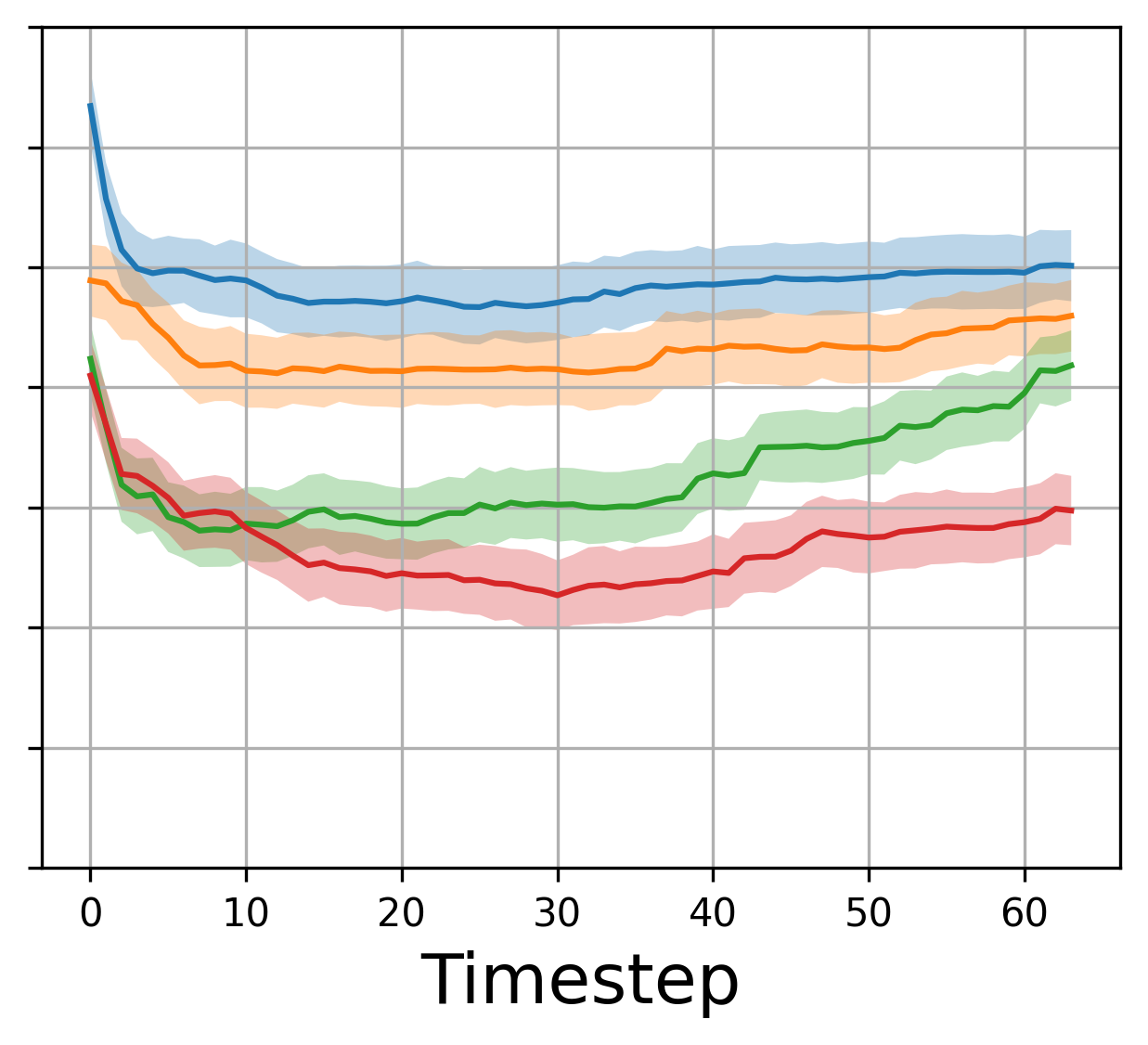}

    (b) $\alpha=0.5$
    \end{minipage}
    \hspace{-30pt}
    \hfill
    \begin{minipage}{0.32\linewidth}
    \centering
    \includegraphics[width=\linewidth]{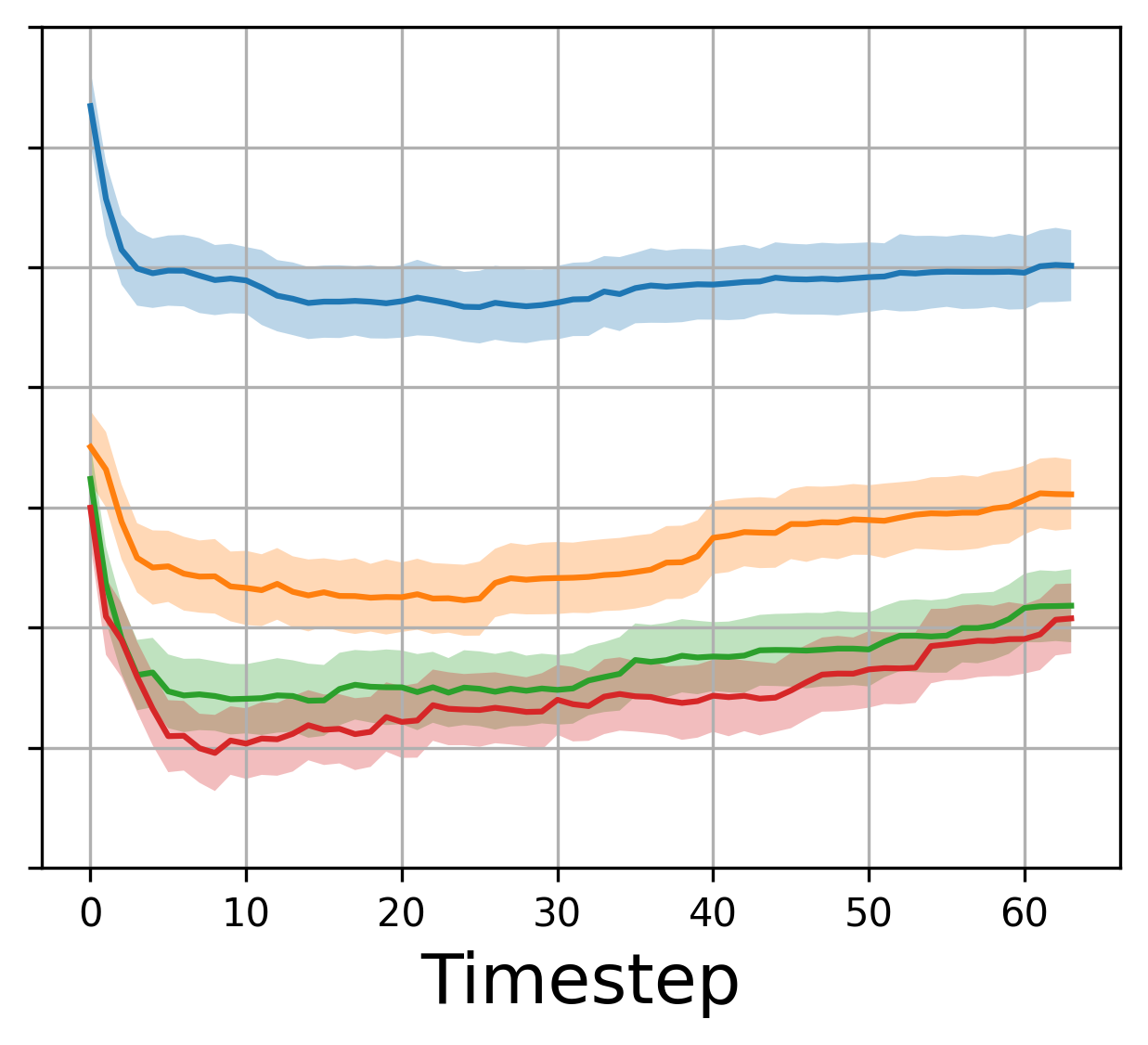}
    
    (c) $\alpha=1.0$
    \end{minipage}
    
    \caption{Entropy of the conditional distribution $p(y|\mathbf{y}_{<t}, \mathbf{x})$ with respect to different $\alpha$ values. Filled regions denote
the standard deviation across training runs according to \autoref{sec:setup}.}
    \label{fig:logp_eos}
    \vspace*{-1em}
\end{figure*}

\label{sec:setup}
We follow the standard practice \citep{Ott_2018} of training our sequence-to-sequence models using FairSeq framework \citep{ott2019fairseq}. We use byte-pair encoding implemented in the fastBPE package \citep{sennrich2016bpe}. We use a transformer architecture for our model and train models on our data from scratch\footnote{Informally, we also tried a pre-trained BART \citep{lewis2019bart} but the results were worse.}.

\paragraph{Model architecture} We use the \texttt{transformer\_iwslt\_de\_en} architecture in FairSeq for experiments. It consists of 6 encoder and decoder layers with 4 self-attention heads followed by feed-forward transformations. Both encoder and decoder use embeddings of size 512 while the input and output embeddings are not shared. Both the encoder and decoder use learned positional embedding. We early-stop training based on the validation performance. Evaluation is done on the test set.

\paragraph{Training} We use Adam optimizer \citep{Kingma15adam} with $\beta_1 = 0.9$ and $\beta_2=0.98$. We use the inverse square root learning scheduler with 4,000 warm-up steps. We use the initial learning rate of $5 \times 10^{-4}$, dropout rate of 0.3~\citep{srivastava14dropout} , and weight decay with its rate set to $10^{-4}$.
We use label smoothing with $0.1$ of probability smoothed uniformly during training.
We modify the training objective \autoref{eq:objective} by adding oversmoothing loss \citep{kulikov2021characterizing} with a coefficient of 0.9 and unlikelihood loss \citep{unlikelihood} with a coefficient of 0.5. All training was performed on VMs with single V100 GPUs, we estimate 200 GPU hours as the total amount required for the completion of this work. 

\paragraph{Early stopping} We use early stopping for model selection based on the value of the objective function computed on the validation set. We evaluate the model on the development set every 2K updates ($\sim$4K tokens per update). We stop training when the objective has not improved over more than 5 consecutive validation runs. It takes approximately 75K updates to an early stop.

\paragraph{Decoding} We use beam search implementation from FairSeq. We decode using the beam size of 5. We set the lower- and upper-bound of a generated output to be, respectively, 0 and $1.2\cdot ||\mathbf{x}|| + 10$. We do not use either length normalization or length penalty since we apply oversmoothing loss.

\paragraph{Lexically constrained decoding baseline} 

Apart from using the unregularized version of the model as a baseline, we compare the proposed approach with the lexically constrained decoding approach \citep{post18dba}. We stick to the \texttt{LexicallyConstrainedBeamSearch} implementation of the Dynamic Beam Allocation (DBA) algorithm that ensures the presence of provided tokens in the generated output. DBA implements an optimized version of the Grid Beam Search \citep{hokamp17gbs}. DBA is training-agnostic and is used only during generation. We apply DBA for the baseline model. Given the non-uniform distribution of utilization rates, for each source we leave only medical concepts $c$ with $r_{\text{id}}(c) > \tau$ for some threshold $\tau$. We report results for $\tau=0.6$, which we select by running an extensive grid search.

\section{Results}
\label{sec:experiments}

\subsection{Effect of knowledge injection during training on model's utilization rate}

\begin{table*}[!ht]
\centering
\begin{tabular}{l|c|lll}                                                                             & $\alpha$ & BERTScore                        & Concept-F1                                      & GPT-2 Perplexity                        \\[1mm] \hline
Baseline                                                                            & 0.0    & 22.48 \scriptsize{$\pm$0.66} & 57.43\scriptsize{$\pm$3.73}                      & \textbf{5.53}\scriptsize{$\pm$0.04}  \\ \hline 
DBA &    -    & 23.59\scriptsize{$\pm$0.28}  & \textbf{79.83}\scriptsize{$\pm$0.43} & 11.96\scriptsize{$\pm$0.05} \\ \hline

\multirow{4}{*}{Unweighted (ours)}                                                               & 0.25   & 25.09\scriptsize{$\pm$0.69}  & 58.19\scriptsize{$\pm$2.11}                      & 5.91\scriptsize{$\pm$0.07}  \\
                                                                                    & 0.5    & 25.42\scriptsize{$\pm$0.56}  & 58.91\scriptsize{$\pm$6.83}                      & 5.65\scriptsize{$\pm$0.03}  \\
                                                                                    & 0.75   & 26.22\scriptsize{$\pm$0.35}  & 60.83\scriptsize{$\pm$5.96}                      & 6.28\scriptsize{$\pm$0.02}  \\
                                                                                    & 1.0    & 26.74\scriptsize{$\pm$0.43}  & 61.05\scriptsize{$\pm$7.48}                      & 6.18\scriptsize{$\pm$0.05}  \\ \hline

\multirow{4}{*}{Concept weighted  (ours)}                                                           & 0.25   & 28.29\scriptsize{$\pm$0.19}  & 60.87\scriptsize{$\pm$3.86}                      & 6.93\scriptsize{$\pm$0.05}  \\
                                                                                    & 0.5    & 28.19\scriptsize{$\pm$0.20}  & 60.36\scriptsize{$\pm$2.03}                       & 8.49\scriptsize{$\pm$0.05}  \\
                                                                                    & 0.75   & 28.08\scriptsize{$\pm$0.15}  & 64.09\scriptsize{$\pm$1.85}                      & 7.95\scriptsize{$\pm$0.080}  \\
                                                                                    & 1.0    & 27.82\scriptsize{$\pm$0.25}  & 63.05\scriptsize{$\pm$2.49}                      & 9.37\scriptsize{$\pm$0.10}  \\ \hline
\multirow{4}{*}{Semantic weighted  (ours)}                                                     & 0.25   & 28.97\scriptsize{$\pm$0.56}  & 69.10\scriptsize{$\pm$2.12}                      & 7.01\scriptsize{$\pm$0.29}  \\
                                                                                    & 0.5    & \textbf{30.54}\scriptsize{$\pm$0.78}  & 74.98\scriptsize{$\pm$3.91}                      & 6.84\scriptsize{$\pm$0.03}  \\
                                                                                    & 0.75   & \textbf{31.48}\scriptsize{$\pm$0.86}  & 75.77\scriptsize{$\pm$3.30}                      & 6.96\scriptsize{$\pm$0.11}  \\
                                                                                    & 1.0    & \textbf{30.59}\scriptsize{$\pm$0.63}  & 75.02\scriptsize{$\pm$2.18}                      & 6.94\scriptsize{$\pm$0.12}  \\
\end{tabular}
\vspace*{1em}
\caption{Automated metrics scores for different model setups. We report average score and standard deviation over five random seeds. We highlight in bold the best average and all scores having overlapped standard deviation intervals with the best score.}
\label{tab:automatic-metrics}
\end{table*}

We evaluate whether the knowledge injection through regularization (\autoref{sec:regularizer})  has the desired effect of improving model estimate of the utilization rate, $r_{\phi}$. Because the test set is too small to effectively estimate per-concept utilization rate, we instead compute it for semantic types.  In \autoref{fig:rel_util} we use semantic relative error (\autoref{eq:relative-error}) to compare models trained with $\alpha \in \{0,0.25,0.5,0.75,1\}$ that either use unweighted loss $l_u$ (which uplifts all medical concepts equally, ``Unweighted") or a weighted loss $l_w$ with the $\phi$ being identity (``Concept weighted'') or mapping concepts to semantic types (``Semantic weighted''). In addition, as a baseline we also compare an unregularized model that uses DBA for generation (``DBA''). For a detailed breakdown of relative errors for each combination see the Supplementary Material. 

\begin{definition}[Semantic relative error]
Relative error for semantic type $s$ computed from $\hat{r}_{\phi}$ estimated from model derived output sequences and $r_{\phi}$ estimated from reference sequences. $c_s$ is any concept for which $\phi(c)=s$ holds and the value of $\epsilon_s$ in not dependent on the choice of $c_s$. 
\begin{align}
\epsilon_s = \frac{\|\hat{r}_{\phi}(c_s)- r_{\phi}(c_s)\|}
{r_{\phi}(c_s)}
\label{eq:relative-error}
\end{align}

\end{definition}

In \autoref{fig:rel_util}a we present the relative error for different $\alpha$ as a function of semantic type frequency in the test set. For each point (a given semantic type and $\alpha$) we take the lowest relative error among \{``Unweighted'', ``Concept weighted'', and ``Semantic weighted''\}.  The highest relative errors are seen for $\alpha=0$, which corresponds to no regularization. For other values of $\alpha$ the difference is not statistically significant, although, for very rare semantic types, $\alpha=0.25$ appears to perform worse than models with higher regularization strength. This shows that our external knowledge informed regularization has a significant impact on a relative error, but the utilization rate estimate is not sensitive to the exact weight of the regularization term.

In \autoref{fig:rel_util}b we present relative error for different training procedures, \{``Unweighted'', ``Concept weighted'', and ``Semantic weighted''\}, as well as a baseline of ``DBA.'' For each point (a given semantic type and training procedure) we choose an $\alpha$ that gives the lowest relative error. We find that ``DBA" baseline, which is a constrained generation procedure applied to an unregularized model, performs worse than any of the regularized models, although it does outperform the unregularized model ($\alpha=0$ in \autoref{fig:rel_util}a). While not significant, we also see that for rare semantic types ``Semantic weighted'' seems to perform the best, which aligns with our expectation that the utilization rate is hard to estimate for very rare concepts.

\subsection{Effect of knowledge injection during training on model's uncertainty} 

We analyze the effect of utilization regularization on the model's uncertainty at every timestep.
Uncertainty at timestep $t$ is defined as an entropy of model's distribution on each timestep $t$ (here $\mathbf{y}_{<t}$ is the decoded sequence up to $t$-th timestep, $y$ is an arbitrary token from the target vocabulary):
\begin{align}
    H_t(\mathbf{y}_{<t}, \mathbf{x}) = -\sum_{y} p(y|\mathbf{y}_{<t}, \mathbf{x}) \log p(y|\mathbf{y}_{<t}, \mathbf{x})
\end{align}

We consider the defined uncertainty on earlier timesteps, where the model's distribution is closer to marginal. As the proposed method pushes up the marginal probability of the medical concepts, we claim that models' uncertainty decreases with the regularization. Moreover, care plan instructions typically introduce crucial concepts at the beginning of an instruction. Thus, we claim that early timesteps uncertainty matters for the precise decoding of instructions.

This is confirmed by \autoref{fig:logp_eos}. We observe that uncertainty drops monotonically as the $\alpha$ weight increases. In particular, uncertainty on early timesteps heavily drops as a result of utilization minimization. Hence, the model becomes more confident in selecting principal concepts at the beginning of an instruction.
In contrast to the baseline, all regularized models' uncertainty start to increase for $t > 10$. As fewer concepts appear in the instruction end, the marginal probability maximization flattens the conditional distribution. However, the uncertainty does not degrade in comparison to the baseline. Thus, the proposed regularization effectively improves the confidence of the model on early timesteps.

\subsection{Results on Care plan instructions task }

\begin{table*}[!ht]
\centering
\begin{tabular}{l|llll}
                         & Relevance                      & Usability                      & Fluency                        & Degeneracies, \%                  \\[1mm] \hline
Baseline                 & 2.50\scriptsize{$\pm$0.12} & 3.18\scriptsize{$\pm$0.27} & \textbf{4.17}\scriptsize{$\pm$0.14} & \textbf{0.10}\scriptsize{$\pm$0.01} \\
DBA                      & 3.36\scriptsize{$\pm$0.15} & 3.35\scriptsize{$\pm$0.16} & 3.91\scriptsize{$\pm$0.18} & 0.21\scriptsize{$\pm$0.05} \\
Unweighted     (ours)           & 3.56\scriptsize{$\pm$0.12} & 3.21\scriptsize{$\pm$0.28} & \textbf{4.26}\scriptsize{$\pm$0.08}  & \textbf{0.10}\scriptsize{$\pm$0.02} \\
Concept weighted   (ours)    & \textbf{3.79}\scriptsize{$\pm$0.06} & 3.72\scriptsize{$\pm$0.05} & \textbf{4.37}\scriptsize{$\pm$0.16} & \textbf{0.12}\scriptsize{$\pm$0.02} \\
Semantic weighted  (ours)& \textbf{3.78}\scriptsize{$\pm$0.14} & \textbf{3.99}\scriptsize{$\pm$0.19} & \textbf{4.42}\scriptsize{$\pm$0.13}  & \textbf{0.12}\scriptsize{$\pm$0.012}\\
\end{tabular}
\caption{Evaluation using medical experts. Fluency, Usability, and Relevance are scored on a scale from 1 to 5. We also report the percentage of premature or repetitive outputs (Degeneracies). We report average score and standard deviation of experts' scores. We highlight in bold the best average and all scores having overlapped standard deviation intervals with the best score.}
\label{tab:human-eval}
\end{table*}

\noindent {\bf Automated evaluation:} The precise and complete concepts utilization directly affects the quality of instruction. We first quantify the quality by calculating automatic metrics to judge the relevance, fluency, and concept utilization rate in comparison to the reference instructions. We use BERTScore \citep{bertscore} to estimate the similarity between reference and candidate, GPT-2 perplexity for \citep{gpt2ppl} to assess the coherence (fluency) of the candidate, and concept overlap \citep{joshi2020dr} to measure the percentage of medical concepts used in both candidate in reference.

\autoref{tab:automatic-metrics} presents the automatic evaluation results. The scores indicate that incorporating knowledge correlates with relevance and concept overlap. We highlight three observations. First, the regularization is effective in terms of quality and concept overlap. We observe significant quality improvement compared to both the baseline and DBA. Moreover, weighted versions of the model outperform the unweighted setup. Thus, injecting more knowledge into the model, such as empirical utilization weights, results in better quality. Second, the impact of the regularization hardly depends on the $\alpha$ weight. Third, the GPT-2 perplexity degrades. This demonstrates that the regularization impacts the model distribution, so the fluency of the model may deteriorate. This trade-off, however, has no negative impact on the quality given the improved BERTScore. For qualitative results, please see the Supplementary Material.

\noindent {\bf Medical experts evaluation:}  To get a more precise medical assessment, we conduct human evaluation with medical experts. We randomly sample 100 dialogues from the test set and generate candidates with each model setup setting $\alpha=1.0$. We ask five doctors to evaluate the relevance to the dialogue, medical usability (if the generated instruction can be used in any care plan), and grammatical correctness (fluency) on a scale from 1 to 5. Additionally, we ask assessors to indicate degenerate generations, i.e., premature or repetitive sequences. Exact questions and interface screenshots can be found in the Supplementary Material.

As shown in \autoref{tab:human-eval}, we claim that both weighted versions achieve significant improvement in relevance and usability, which are target medical metrics. In contrast to the GPT-2 perplexity, medical experts report equal fluency for all models but DBA. We explain this discrepancy with vocabulary shift as GPT-2 is not trained on a healthcare corpus. Finally, utilization rate regularization does not affect the number of degenerate outputs. Hence, the proposed solution effectively induces knowledge in the model distribution without corrupting generated text correctness. This is not true for DBA, which struggles from a lack of coherence and degenerate outputs while producing more relevant and usable instructions.

\section{Conclusion}
In this work, we tackle the problem of under-generation of rare but important tokens in sequence-to-sequence models. We show that external knowledge can be effectively injected into the sequence-to-sequence models and mitigate the problem of lexical precision.  We characterize the problem by identifying a set of low-frequency but important concepts and defining their utilization rate, which estimates the probability of a concept that is present in the source to be also present in the reference. We confirm that modern well-trained sequence-to-sequence models suffer from under-estimating utilization rates, and propose a way to directly maximize it during training. We design a differentiable proxy based on the marginal entropy and propose a regularized training objective. Since some concepts may be omitted from the reference, we extend the approach by applying weights, which restrict the regularization impact of low-utilized concepts or their semantic types. 

We perform a case study in automatic care plan generation from medical dialogues. We experiment with a custom internal dataset and observe the effectiveness of the approach. We also compare a previous approach for external knowledge injection -- dynamic beam allocation (DBA). First, we find  that regularization improves the model's utilization rate by pushing it closer to the empirical values observed in reference sequences. Second, regularization reduces the model's uncertainty at early timesteps: exactly where concepts are typically introduced. Third, we observed a significant (in terms of standard deviations) quality improvement. More specifically, we did a human evaluation of relevance, concept overlap, medical usability, and fluency using five medical experts. The results revealed the enhanced relevance and usability of generated instructions while, unlike DBA, maintaining high fluency and low degeneracy.

\paragraph*{Ethics Statement:} This work was done as part of a quality improvement activity as defined in 45CFR §46.104(d)(4)(iii) -- ``health care operations'' secondary research.
\paragraph*{Reproducibility statement:} Code used for training regularized sequence-to-sequence models in this paper is available at \url{https://github.com/curai/curai-research/tree/main/careplan-charting}. However, data will not be shared due to patient privacy and HIPAA compliance. as it contains significant amount of Patient Health Information (PHI) and cannot be shared.
\paragraph*{Privacy concerns:} Our research aims to utilize knowledge to enhance NLG systems. However, we also acknowledge the privacy concerns associated with leveraging sensitive medical information. All training data was anonymized during preprocessing step, and all personally identifiable information (PII) was removed to protect patient identities in generated outputs. Another privacy consideration is inference leakage, where NLG systems unintentionally reveal sensitive information during generation. We suggest incorporating differential privacy mechanisms to prevent the association of rare tokens or medical concepts with specific individuals.

\section{Limitations}

There are several important limitations to this work that can be split into two categories: (1) method applicability to other domains and (2) method scalability to much larger models.

\paragraph*{Method applicability to other domains.} Utilization rate computation and regularization are possible when there is some external knowledge that can be used to infer which tokens are ``important.'' In particular, our highest-performing model uses token semantic type to compute utilization rates. This limits our approach to sub-domains where there is an external knowledge source that can inform us about important tokens and give us higher-order semantic information about how to group the important tokens. For example, our approach will likely not be very helpful for open-domain conversations.

\paragraph*{Method scalability to much larger models.} We have evaluated our approach for models on the scale of $O(10^8)$ parameters. However, modern state-of-the-art models often involve $O(10^{11})$ parameters, three orders of magnitude larger than models in our experiments. Large language models (LLMs) often still suffer from the under-generation of rare tokens, but our study is insufficient to determine if our approach would still work. We suppose that utilization-rate-based regularization is most likely to be beneficial in the fine-tuning step of LLMs, but verification of this is left for future work.

\bibliography{neurips.bib}
\bibliographystyle{acl_natbib}

\clearpage

\appendix

\section{Semantic relative errors}

\begin{figure*}[!ht]
    \centering
    \includegraphics[width=1\linewidth]{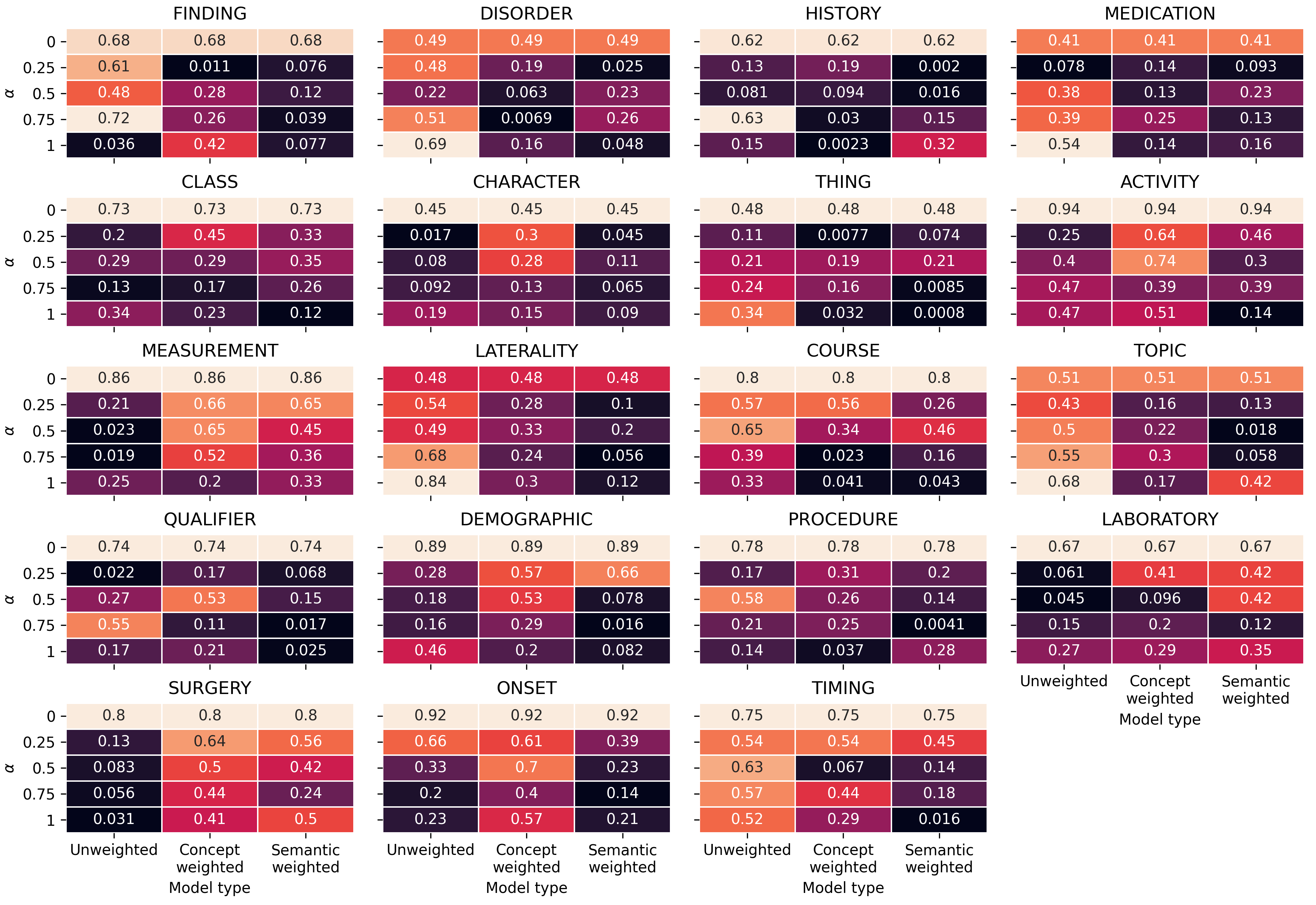}
    \caption{Relative error for each medical semantic type as a function of $\alpha$ and loss type.}
    \label{fig:utilization_rel_rate_full}
\end{figure*}

	\begin{table*}[!ht]
		\footnotesize
		\def\arraystretch{1.4}%
		\hspace*{-0.25cm}
		\begin{tabular}{|p{\linewidth}|}
			\hline
			\multicolumn{1}{|p{\linewidth}|}{\textbf{Instruction}} \\ 
	        We want to evaluate the quality of the automatically generated care plans. In particular, we want to assess the fluency, relevance, clinical usability, and degeneracy of the generated instruction. Given the dialogue with the highlighted prompt (i.e., a span of text that led to instruction), we want to evaluate each property on a scale from 1 to 5. Degenerate instructions stand for extremely short (e.g., “avoid ”), or extremely long “test test test test …”) sequences. There are 4 instruction candidates for each (dialogue, span) pair.  \\  \hline 
		
		\end{tabular}
		\vspace*{4mm}\caption{Instruction provided to the data specialists prior to the human evaluation task submission.}
		\label{tab:instruction}
	\end{table*}

Section 5.1 in the main text discusses the relative error (Equation 7 in the main text) in model computed utilization rate for different semantic types as a function of $\alpha \in \{0,0.25,0.5,0.75,1\}$ and regularization type. The regularizations are $l_u$ (``Unweighted") or a weighted loss $l_w$ with the $\phi$ being identity (``Concept weighted'') or mapping concepts to semantic types (``Semantic weighted''). For $\alpha=0$ all mentioned models are equivalent to the baseline, that does not use any knowledge injection. \autoref{fig:utilization_rel_rate_full} shows the exact values of relative errors for every combination of models.

\section{Human evaluation}

\subsection{Human evaluation UI}

\begin{figure*}[!ht]
    \centering
    \includegraphics[width=1\linewidth]{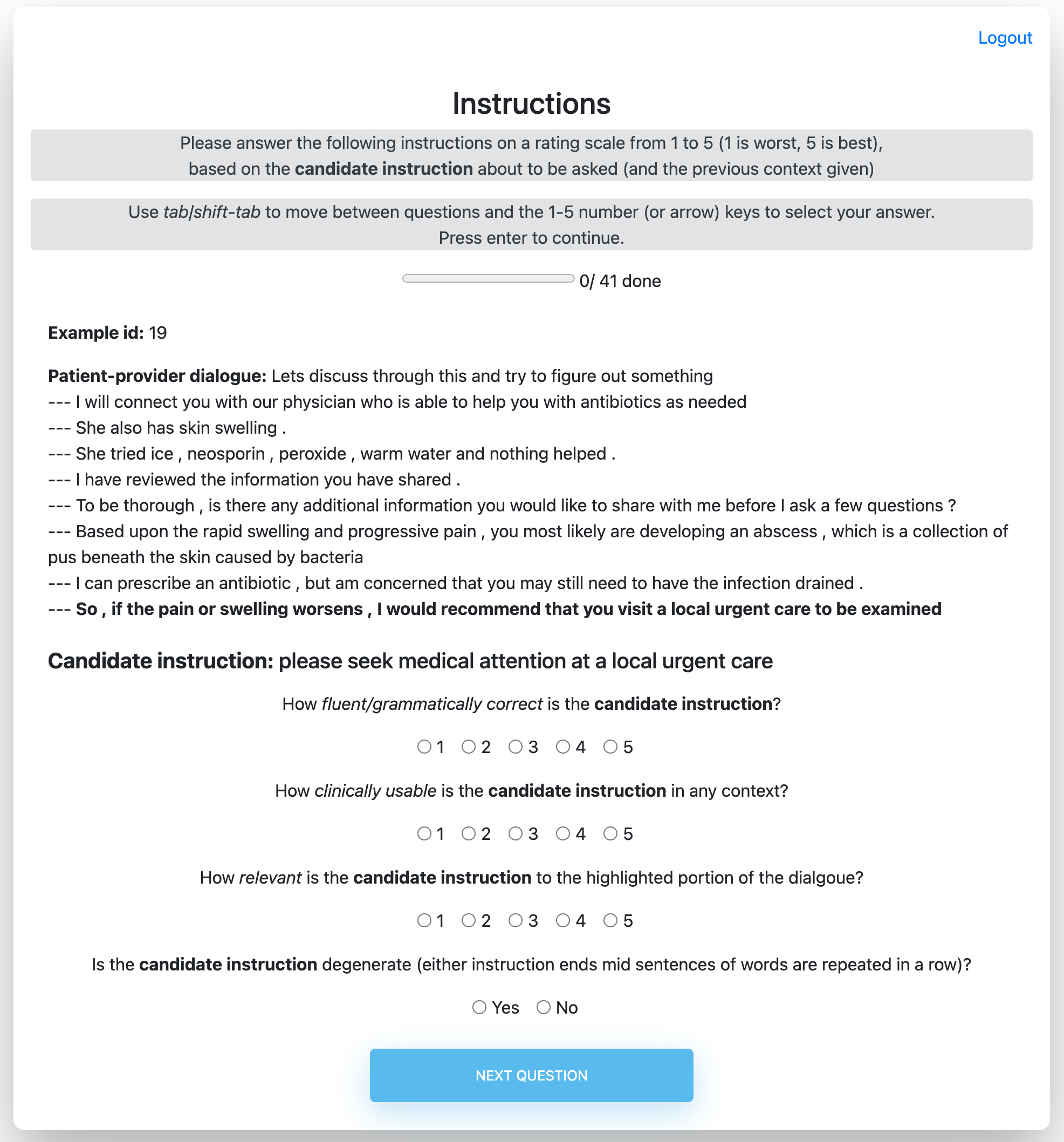}
    \caption{Screen shot of the user interface used in the human evaluation.}
    \label{fig:human_ui}
\end{figure*}

The screen shot of the UI provided to medical experts for evaluation is shown in \autoref{fig:human_ui}.

\subsection{Questions}

We used the following set of questions for medical experts to evaluate every sample:

\begin{enumerate}
    \item \textbf{Usability}: \textit{How clinically usable is the candidate instruction in any context? Please rate on a scale from 1 to 5.}
    \item \textbf{Relevance}: \textit{How relevant is the candidate instruction to the highlighted portion of the dialgoue? Please rate on a scale from 1 to 5.}
    \item \textbf{Fluency}: \textit{How fluent/grammatically correct is the candidate instruction? Please rate on a scale from 1 to 5.}
    \item \textbf{Degeneracies}: \textit{Is the candidate instruction degenerate (either instruction ends mid sentences of words are repeated in a row)? Yes or No.}
\end{enumerate}

\subsection{Evaluation task description}

\autoref{tab:instruction} presents the description of the task that was provided to the medical experts. We also presented it personally to clarify the goals and answer questions.

\section{Qualitative examples }

A complete example of synthezing training samples is given in \autoref{table:complete_example} and qualitative comparison between different models for the final task is in \autoref{tbl:qual}.

\section{Identifying source dialogue turns}

The training data includes only parts of the dialogue relevant to the care plan discussion, which is achieved by the internal segmentation model [work will be published and cited here prior to camera ready]. We then train a FastText model \citepappendix{joulin2016fasttext} on all provided segments. We use spacy framework \citepappendix{spacy2} to split dialogue turns into sentences $\mathbf{x}$ and generate an embedding $E(\mathbf{x})$ for every sentence by averaging the FastText embeddings $e(x_t)$ of the words in a sentence \autoref{eq:embedding}.

\begin{align}
E(\mathbf{x}) = \frac{1}{\|\mathbf{x}\|} \sum_{t=1}^{\|\mathbf{x}\|} e(x_t)
\label{eq:embedding}
\end{align}

We repeat the procedure for the true care plan instructions $\mathbf{y}$. Next, we use a cosine similarity $c$ (\autoref{eq:cosine}) between FastText embeddings of $\mathbf{x}$ and $\mathbf{y}$ with a threshold of 0.85 to map a sentence to the relevant care plan instruction. We omit the unmapped sentences and care plan instructions from the dataset. 

\begin{align}
c(\mathbf{x}, \mathbf{y}) = \frac{E(\mathbf{x}) \cdot E(\mathbf{y})}{\|E(\mathbf{x})\| \|E(\mathbf{y})\|}
\label{eq:cosine}
\end{align}

To improve computational efficiency, we utilize the FAISS framework for mapping \citepappendix{johnson2019billion}.

	\begin{table*}[!ht]
		\footnotesize
		\def\arraystretch{1.4}%
		\hspace*{-0.25cm}
		\begin{tabular}{|p{\linewidth}|}
			\hline
			\multicolumn{1}{|p{\linewidth}|}{\textbf{Patient-Provider conversation. Shown only  provider turns for brevity}} \\
\textbf{MD:} Based on your symptoms, it sounds like you have an upper respiratory infection. \\ 
\textbf{MD:} \textcolor{purple}{For the sore throat and any cough, you can try OTC cough medicine, but in experience it is not any more effective than home remedies.} \textbf{(1)} \\
\textbf{MD:} A humidifier, or simply breathing in steam like in the shower will help with any chest congestion. \\
\textbf{MD:} \textcolor{blue}{I also recommend gargling with warm salt water, that will help with the throat inflammation.} \textbf{(2)} \\
\textbf{MD:} If you develop severe shortness of breath, you should go to the ER right away \\
\textbf{MD:} Tonsillitis is inflammation and possibly infection of your tonsils. \\
\textbf{MD:} Yes, I generally recommend giving it a week, and during that time continue to gargle with warm salt water, taking motrin and tylenol as needed for pain, \textcolor{orange}{drinking/eating soft food so it doesnt irritate your throat} \textbf{(3)} \\
\textbf{MD:} If your tonsils are getting larger and more painful, or you are having severe pain with swallowing , please let us know and we will re-assess \\
\textbf{MD:} Upper respiratory infections and throat infections, including tonsillitis, usually go away in 1-2 weeks, but if its lasting longer than that please let us know. \\
\textbf{MD:} \textcolor{blue}{Please do gargle with the warm salt water as discussed, that will help the swelling more.} \textbf{(2)} \\
\textbf{MD:} \textcolor{teal}{One more recommendation is to try TheraFlu cold and cough - its available over the counter - and will help with pain and congestion as well.}  \textbf{(4)} \\
\textbf{MD:} Please feel free to reach out to us with further questions at any time.  \\ \hline 
 \textbf{True care plan instructions} \\
\textbf{(1)}: Medication Plan: \textcolor{purple}{Take Ibuprofen or Tylenol as needed, as directed, for pain.} \\
\textbf{(2)}: Instruction: \textcolor{blue}{Gargle with warm salt water several times a day to help throat inflammation.} \\
\textbf{(3)}: Instruction: \textcolor{orange}{Avoid any harsh or irritating foods that may worsen or further irritate your sore throat.} \\
\textbf{(4)}: Medication Plan: \textcolor{teal}{Take TheraFlu Cold and Cough, available over the counter, as needed, as directed, for pain and congestion.} \\
\hline
\textbf{Concepts with semantic types} \\
\textbf{(1)}: \textcolor{purple}{sore throat (FINDING), cough (DISORDER)} \\
\textbf{(2)}: \textcolor{blue}{water (FINDING), throat inflammation (FINDING), swelling (CLASS)} \\
\textbf{(3)}: \textcolor{orange}{drinking (FINDING)} \\
\textbf{(4)}: \textcolor{teal}{cough (DISORDER), TheraFlu (MEDICATION), pain (CLASS), congestion (CLASS)} 
 \\ \hline 
		\end{tabular}
		\vspace*{2mm}\caption{Training examples construction. The only inputs are the provider turns of the care plan section of the dialog (first box) and the care plan instructions (last box) from the EHR. For each care plan instruction, we identify the pertinent dialog turn (see the main paper for the details) and use those turns as the source for the instructions. We have marked these mappings in the diagram using numbering them from (1) to (4). Notice how instruction (2) maps to more than one dialog turn in the conversation. We use these mappings to create training pairs. Figure also shows medical concepts along with their semantic types in the box labeled `Concepts with semantic types'}
		\label{table:complete_example}
	\end{table*}

\clearpage

\begin{table*}[]
\small
\resizebox{1\linewidth}{!}{
\begin{tabular}{@{}p{0.25\linewidth}p{0.25\linewidth}@{\hskip 0.3in}p{0.6\linewidth}@{}}
\toprule
\multicolumn{2}{c}{\textbf{Test data}} & \multicolumn{1}{c}{\textbf{Generations}} \\ \midrule
\multicolumn{1}{c}{Source} & \multicolumn{1}{c}{Reference} & \multicolumn{1}{c}{Generated Instructions} \\ \midrule
\begin{tabular}[t]{p{\linewidth}}
I recommend you to take claritin daily for a few days and add flonase spray to decrease the pressure in the upper airway and your ear \end{tabular} & \begin{tabular}[t]{p{\linewidth}} Over the counter anti-allergic medications like Claritin and Flonase spray may be helpful. \end{tabular} & 
\begin{tabular}[t]{@{}p{\linewidth}@{}}
B: \color{purple}\textit{Recommend claritin days add flonase spray decreased pressure upper airway ear}\\ \\
DBA: \color{orange} \textit{Recommend claritin and flonase}\\ \\
U (ours): \color{blue} \textit{Recommend claritin and flonase spray to decrease pressure}\\ \\
CW (ours): \color{teal} \textit{Recommend claritin and flonase spray to decrease pressure} \\ \\
SW (ours): \color{violet} \textit{Recommend claritin and flonase spray to decrease pressure} \end{tabular} \\ \midrule

\begin{tabular}[t]{p{\linewidth}} Continue taking your Vienva every day for now, and get tested for STD and pregnancy \end{tabular} & \begin{tabular}[t]{p{\linewidth}} Get tested for STD and pregnancy. You can do this with your current primary care doctor. \end{tabular} &
\begin{tabular}[t]{@{}p{\linewidth}@{}}
B: \color{purple} \textit{Undergo std testing and women health exam} \\ \\
DBA: \color{orange} \textit{Continue taking your vienva every day} \\ \\
U (ours): \color{blue} \textit{Undergo std and pregnancy test} \\ \\
CW (ours): \color{teal} \textit{Undergo std and pregnancy test} \\ \\
SW (ours): \color{violet} \textit{Continue taking every day} 
\end{tabular}\\ \midrule

\begin{tabular}[t]{p{\linewidth}}In the meantime, try to eat bananas and drink citrus products to add potassium to your diet \end{tabular}& \begin{tabular}[t]{p{\linewidth}}Eat potassium-rich foods \end{tabular}& 
\begin{tabular}[t]{@{}p{\linewidth}@{}}
B: \color{purple} \textit{Continue the health diet} \\ \\
DBA: \color{orange} \textit{You will require repeat labs to check your potassium level again} \\ \\
U (ours): \color{blue} \textit{Continue the health diet} \\ \\
CW (ours): \color{teal} \textit{You will require repeat labs to check your potassium level again} \\ \\
SW (ours): \color{violet} \textit{You will require repeat labs to check your potassium level again} \end{tabular}\\ \midrule

\begin{tabular}[t]{p{\linewidth}}There is an antibiotic called Doxycycline which I can prescribe to cure the infection \end{tabular} & \begin{tabular}[t]{p{\linewidth}} Doxycycline 100 mg oral tablet has been prescribed for you. \end{tabular}& 
\begin{tabular}[t]{@{}p{\linewidth}@{}}
B: \color{purple} \textit{Take antibiotics as} \\ \\
DBA: \color{orange} \textit{Doxycycline} \\ \\
U (ours): \color{blue} \textit{Take doxycycline as needed} \\ \\
CW (ours): \color{teal} \textit{Take doxycycline as directed to cure}  \\ \\
SW (ours): \color{violet} \textit{Doxycycline has been prescribed for you}
\end{tabular}\\ \bottomrule
\end{tabular}}
\caption{Qualitative examples from the test set comparing different methods. We use different color and abbreviations for each method: \textcolor{purple}{B} for \textcolor{purple}{B}aseline, \textcolor{orange}{DBA} for \textcolor{orange}{D}ynamic \textcolor{orange}{B}eam \textcolor{orange}{A}llocation, \textcolor{blue}{U} for \textcolor{blue}{U}nweighted, \textcolor{teal}{CW} for \textcolor{teal}{C}oncept-\textcolor{teal}{W}eighted, and \textcolor{violet}{SW} for \textcolor{violet}{S}emantic-\textcolor{violet}{W}eighted. In each block, we present a source dialog turn (source), and the reference care plan instruction for that turn (reference). In the last column, we show the generated care plan instruction for the source by the different methods. You can see how our final model (semantic weights) provides more detailed instructions including capturing medical concepts correctly.}
\label{tbl:qual}
\end{table*}

\clearpage
\bibliographyappendix{appendix.bib}
\bibliographystyleappendix{acl_natbib}

\counterwithin{figure}{section}
\counterwithin{table}{section}

\clearpage


\end{document}